\documentclass[10pt,twocolumn,letterpaper]{article}

\usepackage{cvpr}              

\usepackage{multirow}
\usepackage{algorithm}
\usepackage{algpseudocode}
\usepackage{subcaption}

\definecolor{cvprblue}{rgb}{0.21,0.49,0.74}
\usepackage[pagebackref,breaklinks,colorlinks,allcolors=cvprblue]{hyperref}

\title{Adaptive Temperature Based on Logits Correlation in Knowledge Distillation}

\author{Kazuhiro Matsuyama$^{1}$ \and Usman Anjum$^{1}$ Satoko Matsuyama$^{2}$ \and Tetsuo Shoda$^{2}$ \and Justin Zhan$^{1}$\\
$^{1}$University of Cincinnati, Department of Computer Science\\
$^{2}$Cincinnati Children's Hospital Medical Center\\
}

\begin{document}
\maketitle
\begin{abstract}
Knowledge distillation is a technique to imitate a performance that a deep learning model has, but reduce the size on another model. It applies the outputs of a model to train another model having comparable accuracy. These two distinct models are similar to the way information is delivered in human society, with one acting as the "teacher" and the other as the "student". Softmax plays a role in comparing logits generated by models with each other by converting probability distributions. It delivers the logits of a teacher to a student with compression through a parameter named temperature. Tuning this variable reinforces the distillation performance. Although only this parameter helps with the interaction of logits, it is not clear how temperatures promote information transfer. In this paper, we propose a novel approach to calculate the temperature. Our method only refers to the maximum logit generated by a teacher model, which reduces computational time against state-of-the-art methods. Our method shows a promising result in different student and teacher models on a standard benchmark dataset. Algorithms using temperature can obtain the improvement by plugging in this dynamic approach. Furthermore, the approximation of the distillation process converges to a correlation of logits by both models. This reinforces the previous argument that the distillation conveys the relevance of logits. We report that this approximating algorithm yields a higher temperature compared to the commonly used static values in testing.
\end{abstract}
\section{Introduction}
\label{sec:intro}

The advancement of deep neural network has much success to solve tasks in several fields~\cite{DBLP:journals/corr/abs-2308-04268}. However, there is a problem requiring a huge training overhead and computational time, which makes it difficult to implement to implement them to edge devices or real-time systems. Knowledge Distillation (KD)~\cite{DBLP:journals/corr/HintonVD15} is a remarkable option to solve the problem and recognized by both academia and industry~\cite{10.1609/aaai.v37i2.25236}. KD applies the outputs of a model to train another model having comparable accuracy. These two distinct models are similar to the way information is delivered in human society, with one acting as the "teacher" and the other as the "student". For instance, Hinton \textit{et al.}~\cite{DBLP:journals/corr/HintonVD15} suggest a method for minimizing Kullback–Leibler divergence (KL divergence)~\cite{DBLP:journals/corr/abs-2404-00936} between the predicted values of both models.

There are two major approaches called logit-based and feature-based. The former improve a performance of a student model by only using the predicted values called logit. The advantage is to train a new model without gathering internal confidential information of a referenced model. Feature-based approaches allow all access to the model and is likely to obtain a better performance than logit-based methods by reaching outputs of intermediate layers. Given its attributes these are known as a white-box and a black-box approaches~\cite{DBLP:journals/corr/abs-2407-01885}. These approaches intrinsically require enabling the comparison of outputs from both the teacher and the student. 

The temperature $\tau$ plays the important role during the process. It makes a predicted probability distribution smooth and this has the effect of reinforce other information than the correct label. The different parameter setting makes the distillation pay attention to logit shift~\cite{DBLP:conf/icml/ChandrasegaranT22, DBLP:journals/corr/abs-2202-07940}. Intuitively, the sample-wise temperature should be effective for a better distillation, but a static value was commonly accepted. Several methods~\cite{10.1609/aaai.v37i2.25236, zheng2024knowledge} proposes methods to calculate a dynamic temperature. Although the adaptive parameter works in the context of soft labels with softmax, previous works rarely mentioned the approximation.

In this paper, we propose that a maximum logit for each sample determines the shared temperature by Taylor series approximation~\cite{DBLP:journals/corr/BrebissonV15}. The approximation function that takes z-score normalization~\cite{Sun_2024_CVPR} into account shows that when the temperature is one or higher, the information is aggregated into the lower-order terms. This is satisfied in the case of image classification tasks that follow a single-peak distribution. In addition, according to the approximation conditions, the lowest temperature maximizes the KL Divergence. This low-order term is equivalent to the correlation coefficient. Our contributions are follows:
\begin{itemize}
    \item We propose to a novel method to calculate a sample-wise temperature based on the maximum logit.
    \item We derive $n^{\text{th}}$ order Taylor Series approximation converges the correlation of logits.
    \item Our method improves existing KD methods using a temperature on a benchmark dataset.
\end{itemize}
\section{Related Work}
\label{sec:related_work}

Knowledge distillation~\cite{DBLP:journals/corr/HintonVD15} applies the outputs of a model to train another model having comparable accuracy. These two distinct models imitate transferring information in human society, calling one the teacher and the other the student. This derives from that a student model refers the output of the teacher model to obtain a better performance. Following the different type of the output used during the transfer, there are two approaches as logit-based~\cite{DBLP:journals/corr/HintonVD15, Zhao_2022_CVPR, 10.1609/aaai.v37i2.25236, zheng2024knowledge, Sun_2024_CVPR} and feature-based~\cite{DBLP:journals/corr/RomeroBKCGB14, park2019relational, DBLP:conf/iclr/TianKI20, DBLP:conf/cvpr/Chen0ZJ21, DBLP:conf/cvpr/ChenMZWF022, DBLP:conf/cvpr/GuoYL023}.

Logit-based approach only transfers the logits to a student model. For instance, KD loss~\cite{DBLP:journals/corr/HintonVD15} utilizes Kullback–Leibler divergence (KL divergence), comparing the logits generated by both teacher and student models. The student is able to generate the similar output of the teacher by minimizing the loss function. This comparison occurs with compressing logits through a temperature. The parameter makes the probability distribution smooth after applying softmax into the compressed logits and emphasize the difference between models.

The temperature parameter controls how much attention students pay to negative logits compared to the average~\cite{DBLP:conf/icml/ChandrasegaranT22, DBLP:journals/corr/HintonVD15, Jin_2023_CVPR}. When the temperature is very high, the logits distribution becomes uniform, and the between the compressed logits of the two models converge. In contrast, the low temperature emphasizes the difference between logits. The optimal temperature has been determined by trial and error, but recently there has been a shift towards dynamic control. Liu \textit{et al.}~\cite{liu2022metaknowledgedistillation} propose a method to update a temperature by meta-learning. Zheng \textit{et al.}~\cite{10.1609/aaai.v37i2.25236} develop a curriculum-based method that gradually changes the distillation from simple to difficult knowledge. The combination of the adversarial learning framework and the temperature increase schedule enabled temperature changes that matched the improvement in the accuracy of the student model. Kaixiang \textit{et al.}~\cite{zheng2024knowledge}, was inspired by R\'enyi Entropy and redefined temperature by only providing a temperature setting to the teacher model. The temperature is substituted by a power transformation of probabilities. This makes the probability distribution of the student model, obtaining a smooth distribution. They propose the sample-wise implementation with weighted values. These methods reported improved performance over the conventional global temperature setting, suggesting that dynamic temperature settings promote knowledge distillation. Sun \textit{et al.}~\cite{Sun_2024_CVPR} showed the irrelevance between the temperatures of different samples using the Lagrange multiplier and that the issue of the shared temperatures is solved by z-score standardization. The temperature derived from this multiplier provides different temperatures for various samples.

The adaptive temperature can treat it as the approximation of the probabilities. The logits by a student get close to ones of a teacher by the parameter. Several works explored alternatives to softmax used in KL divergence. Taylor Softmax~\cite{DBLP:conf/nips/VincentBB15} is an alternative method to softmax that uses Taylor expansion to approximate functions. Taylor expansion expresses a function as the sum of n derivatives. This approximation is valid within a certain range of neighboring points. SM-Taylor softmax~\cite{DBLP:conf/delta/0001CGVHM21} is a generalization of Taylor softmax and introduces soft margins. The additivity of Taylor expansion has desirable properties for matrix operations. In this study, we approximate the softmax and log softmax functions and show that the approximation converges to the correlation between logits according to temperature scaling.

\section{Methodology}
\label{sec:Methodology}

We show that the comparison of z-score standardized datasets converges the correlation by approximating KL divergence. A shared temperature compresses the logits of the teacher and student models by the same value and converts them to probability distributions through softmax. Subsection~\ref{subsec:taylor_series_approx} approximates softmax for logits scaled by $\tau$. KL divergence with the converted function derives additional approximation. The denominator gets either a temperature or the number of sample. This results in the convergence of the lower-order term, which represents the correlation coefficient. In subsection~\ref{subsec:radius_of_convergence}, we derive the radius of the convergence to make this formula work. 

\subsection{Preliminaries}
\textbf{Notation} Given that a dataset $\mathcal{D}$ containing total $N$ samples $\{x_n, y_n\}^K_{n=1}$, $x_n \in \mathbb{R}^{H \times W}$ and $y_n \in [1, N]$ are the $K^{\text{th}}$ input images and the correct label, respectively. $H$, $W$ and $N$ notates the height and the width of the input and the number of classes. A model generates the logit $v_i$ against the input $\mathcal{D}$. Applying z-score standardization $v_i$ yields $z_i$ following $\sum z_i = 0$. For the convenience, the normalized values divided by a scaling parameter $\tau$ called by temperature notates $t_i$. The logits predicted by a teacher model and a student model describes $t^p_i$ and $t^q_i$, respectively.
The conversion of logits to class probabilities archives with the softmax function. The $i^{\text{th}}$ sample's probabilities are
\begin{align}
p_i = \frac{e^{t^p_i}}{\sum_{k=0}^{N-1} e^{t^p_k}}, q_i = \frac{e^{t^q_i}}{\sum_{k=0}^{N-1} e^{t^q_k}}
\end{align}

where $t^p_i$ and $t^q_i$ are logits scaled by $\tau$.

\textbf{Knowledge Distillation} A temperature manages the smoothness of transfering from a teacher model to a student model. When $\tau = 1$, the output obtains a comparison with the logit $z^p_i$ and $z^q_i$. The purpose of this hyper-parameter is the promotion of transferring the relationship of logits in a sample~\cite{Sun_2024_CVPR}. The dynamic change implies to the accuracy improvement~\cite{10.1609/aaai.v37i2.25236}. Here, we prove that KL divergence converges the correlation of logits by the approximation of knowledge distillation loss. The loss follows
\begin{align}
D_{KL}(p||q) = \frac{1}{N} \sum_{i=0}^{N-1} p_i \log(\frac{p_i}{q_i})
\end{align}

where $p$ and $q$ are probability distributions applied softmax into logits divided by $\tau$.

\subsection{Taylor Series Approximation}
\label{subsec:taylor_series_approx}
Taylor series approximates a given function $f$ using $n^{\text{th}}$ order derivatives and their summation~\cite{DBLP:conf/delta/0001CGVHM21}. The approximation of $e^z$ follows
\begin{align}
e^{z} \approx f^n(z) = \sum_{i=0}^{n} \frac{z^i}{i!}
\end{align}

Softmax can converge
\begin{align}
softmax(z_i) &\approx \frac{f^n(z_i)}{\sum_{k=0}^{N-1} f^n(z_k)} \\
\end{align}

where $z_i$ is logits scaled by $\tau$, $i$ and $k$ notates each class for $i,k \in K$ and $N$ is the number of classes $N \in \mathbb{N}$. Applying the $n^{\text{th}}$ order approximation expresses KL divergence as follows

\begin{align}
D_{KL}(p||q) &= \frac{1}{N} \sum_{i=0}^{N-1} p_i \log(\frac{p_i}{q_i}) \\
&\approx \frac{1}{N} \sum_{i=0}^{N-1} \frac{f^n(z^p_i)}{\sum_{k=0}^{N-1} f^n(z^p_k)} \\
& (\log \frac{f^n(z^p_i)}{f^n(z^q_i)} - \log \frac{\sum_{k=0}^{N-1} f^n(z^p_k)}{\sum_{k=0}^{N-1} f^n(z^q_k)}) \notag
\end{align}

Here, logarithm of $f^n(z_i)$ and $\sum f^n(z_k) $ can represent the likeness
\begin{align}
\log (1 + (f^n(z_i) - 1)) &\approx \sum_{i=0}^{m} (-1)^i \frac{(f^n(z_i) - 1)^{i+1}}{i+1}
\end{align}
\begin{align}
\log (1 + (\sum_{k=0}^{N-1} f^n(z_k) - 1) &\approx \sum_{i=0}^{m} (-1)^i \frac{(\sum_{k=0}^{N-1} f^n(z_k) - 1)^{i+1}}{i+1}
\end{align}

Given that $z^p_i = \frac{p_i}{\tau}$, $z^q_i = \frac{q_i}{\tau}$, the denominator is either $N$ or $\tau$, irrespective of $n$. For example, approximating with $n = m = 1$ describes as follows:
\begin{align}
D_{KL}(p||q) &\approx \frac{1}{N} \sum_{i=0}^{N-1} \frac{1 + z^p_i}{N} (z^p_i - z^q_i) \\
&= \frac{1}{N} \sum_{i=0}^{N-1} \frac{z^p_i + z^p_i z^q_i}{N} \label{eq:eq_01} \\
&= \frac{1}{N^2} \sum_{i=0}^{N-1} z^p_i z^q_i
\end{align}
This is satisfied with $n > 2$. The approximated formula consists of terms differentiated $n$ and $m$ times, Eq.~\ref{eq:eq_01} scaled by $\tau$, the denominator of $z_i$. From $N > 2$, This converges the lower term, $\sum z^p_i z^q_i$. In z-score standardized dataset, this is equivalent to correlation coefficient.

\subsection{Radius of Convergence}
\label{subsec:radius_of_convergence}
The updated radius of convergence $R$ depends on the maximum logit. The error term of $\frac{z_i}{\tau}=0$ can describe as $R = 1$, 
\begin{align}
-1 < \frac{z_i}{\tau} < 1 \\
z_i < |\tau| \label{eq:eq_02}
\end{align}
Next, We derive the radius of convergence in the approximation for log softmax. Using $2^{\text{nd}}$ order, the relationship between the summation of the logits and $\tau$ is
\begin{align}
-1 < \frac{z_i}{\tau} + \frac{z_i^2}{2\tau^2} < 1 \\
-2 < \frac{z_i^2}{\tau^2} + 2\frac{z_i}{\tau} < 2
\end{align}
Here, assuming $ \tau > 0$, the range of $\tau$ is
\begin{align}
\frac{z_i^2}{\tau^2} + 2\frac{z_i}{\tau} -2 < 0 \\
2\tau^2 - 2z_i\tau - z_i^2 > 0 \\
\tau < \frac{z_i(1 - \sqrt{3})}{2}, \frac{z_i(1 + \sqrt{3})}{2} < \tau
\end{align}
This satisfies inequality Eq.~\ref{eq:eq_02} and converges the following:
\begin{align}
\frac{max(z_i)(1 + \sqrt{3})}{2} < \tau
\end{align}

\subsection{Truncated Terms}
$2^{\text{nd}}$ order Taylor series approximation in a classification task takes lower remainder than the upper bound. If all logits have either $-\sqrt{3} - 1$ or $\sqrt{3} + 1$, the remainder gets the maximum. However, A single modal distribution consists of a single higher logit and other lower scores. This property implies that the average divergence loss is lower than the bound.


\subsection{Maximum Logit as Temperature}
We implement an algorithm to adjust a temperature from the maximum logit with the $2^{\text{nd}}$ order approximation. Algo.~\ref{alg:alg02} converts the predicted values to z-score standardized data~\cite{Sun_2024_CVPR} for a comparison. Algo.~\ref {alg:alg01} extracts the sample-wise maximum logits from the dataset and multiply the lower bound obtained by the approximation. The imputation to KL divergence through softmax makes the distillation loss.

\begin{algorithm}
\caption{Maximum Logit as Temperature Function.}
\label{alg:alg01}
\textbf{Input:} Input vector $x_t$, $x_s$, the number of terms use in Taylor series, $a$ \\
\textbf{Output:} Temperature $\mathcal{T}(v_t; v_s)$ \\
\textbf{Note:} The function $f^a(\tau)$ depends on the input $a$, conditions given by Taylor approximation.
\begin{algorithmic}[1]
\State $m_t \leftarrow \max(v_t)$
\State $m_s \leftarrow \max(v_s)$
\State $m_\tau \leftarrow \max(m_t, m_s)$
\State \Return $f^a(m_\tau)$
\end{algorithmic}
\end{algorithm}

\begin{algorithm}
\caption{Weighted $\mathcal{Z}$-score Function without Temperature.}
\label{alg:alg02}
\textbf{Input:} Input vector $v$ \\
\textbf{Output:} Standardized vector $\mathcal{Z}(v)$
\begin{algorithmic}[1]
\State $\overline{v} \leftarrow \frac{1}{K} \sum_{k=1}^K v^{(k)}$
\State $\sigma(v) \leftarrow \sqrt{\frac{1}{K} \sum_{k=1}^K \left( v^{(k)} - \overline{v} \right)^2}$
\State \Return $(v - \overline{v}) / \sigma(v)$
\end{algorithmic}
\end{algorithm}

\begin{algorithm}
\caption{Maximum Logits as Temperatures in Knowledge Distillation.}
\label{alg:alg03}
\textbf{Input:} dataset $\mathcal{D}$ with image-label sample pair $\{x_n, y_n\}_{n=1}^N$, Teacher $f_T$, Student $f_S$, Loss $\mathcal{L}_{\text{KD}}$, hyper-parameter for loss $\lambda$ and $\mathcal{Z}$-score function $\mathcal{Z}$ in Algo.~\ref{alg:alg02} \\
\textbf{Output:} Trained student model $f_S$
\begin{algorithmic}[1]
\For{each $(x_i, y_i)$ in $\mathcal{D}$}
    \State $v_i \leftarrow f_T(x_i), \, z_i \leftarrow f_S(x_i)$
    \State $z(v_i) \leftarrow \mathcal{Z}(v_i)$
    \State $z(z_i) \leftarrow \mathcal{Z}(z_i)$
    \State $\tau_i \leftarrow \mathcal{T}(z(v_i); z(z_i))$
    \State $q(v_i) \leftarrow \text{softmax}(z(v_i) / \tau_i)$
    \State $q(z_i) \leftarrow \text{softmax}(z(z_i) / \tau_i)$
    \State $p(z_i) \leftarrow \text{softmax}(z_i)$
    \State Update $f_S$ towards minimizing
    \State \hspace{1em} $\lambda_{\text{CE}} \mathcal{L}_{\text{CE}}(y_i, p(z_i)) + \lambda_{\text{KD}} \tau_i^2 \mathcal{L}_{\text{KD}}((q(v_i), q(z_i)))$
\EndFor
\end{algorithmic}
\end{algorithm}
\section{Experiments}
\label{sec:Experiments}

\textbf{Datasets} We execute experiments on CIFAR-100~\cite{krizhevsky2009learning}. CIFAR-100 is widely accepted dataset to evaluate a performance of a model. The dataset has 60,000 images with $32 \times 32 size$. 50,000 training sets and 10,000 validation sets are prepared. The number of classes is 100, which is separated to 5 super classes consisting of several subclasses. 

\textbf{Baselines}  We evaluate the performance differences of our approach as pre-processing to various logit-based KD approaches with the other modules dealing with temperature changes. Hinton \textit{et al.}~\cite{DBLP:journals/corr/HintonVD15} proposes KD loss with KL divergence scaled by $\tau$. DKD splits KD into the two modules of target class and non-target class. They share a same temperature during training~\cite{Zhao_2022_CVPR}. CTKD is an easy-to-plug-in technique to change a temperature following a pre-defined coefficient per epoch~\cite{10.1609/aaai.v37i2.25236}. TTM and WTTM  apply scaling based on  R\'enyi Entropy instead of a temperature~\cite{zheng2024knowledge}. Our method replaces KD and DKD with the dynamic update and compare the performance against those using CTKD. Furthermore, we conduct feature-based approaches. Romero \textit{et al.}~\cite{DBLP:journals/corr/RomeroBKCGB14} propose FitNet matching intermediate layers to boost a performance. RKD distills structural relations following a similarity of outputs~\cite{park2019relational}. CRD utilizes contrastive learning to capture structures of feature representations~\cite{DBLP:conf/iclr/TianKI20}. ReviewKD archives the distillation by assigning the intermediate information from various layers to a single layer~\cite{DBLP:conf/cvpr/Chen0ZJ21}. SimKD aligns the size of feature maps from hidden units and reuse the pre-trained classifier~\cite{DBLP:conf/cvpr/ChenMZWF022}. CAT-KD aims at matching attentions on each feature representations for both models~\cite{DBLP:conf/cvpr/GuoYL023}.

\textbf{Implementation Details} The training configuration is designed to enable a concrete comparison of different methods. We utilize one GPU for all test scenarios. The batch size is set to 64 over a total of 240 epochs. The learning rate starts at 0.05, except when using the student models MobileNetV2 or ShuffleNetV2, for which it is 0.01. The learning rate decays by a factor of 0.1 at epochs 150, 180, and 210. A weight decay of 0.0005 and a momentum of 0.9 are applied. Other settings can be found on Tab.~\ref{tab:hyperparameter_settings} and~\href{https://github.com/kei813121/atkd}{Github}, All tests examine one time on CIFAR-100.

\subsection{Results}
\textbf{Performance} We evaluate our method and others with various combinations of a teacher and a student model in Tab.~\ref{tab:test1_diff_arch} and Tab.~\ref{tab:test2_same_arch}. Tab.~\ref{tab:test1_diff_arch} shows the top-1 accuracy on CIFAR-100 with the different architecture of a teacher and a student. Tab.~\ref{tab:test2_same_arch} indicates the same test for Tab. \ref{tab:test1_diff_arch} but the same architecture. We compare the different KD approach for a temperature. Our method improves the performance in some models. Evaluating the accuracy with the different architecture gets higher accuracies than existing methods for some patterns. Our adaptive temperature help the vanilla KD~\cite{DBLP:journals/corr/HintonVD15} get a better performance. In most cases of Tab.~\ref{tab:test1_diff_arch}, our method overcomes the accuracy of other modules to calculate a temperature. The similar improvement is confirmed in DKD~\cite{Zhao_2022_CVPR} after applying our method.

\textbf{Limitation} The main result with CIFAR-100 implies that our method has a trouble when the logit distribution doesn't follow a single modality. As shown in subsection~\ref{subsec:radius_of_convergence}, our method assumes that the soft labels have a strong peak. Utilizing the maximum value as a temperature yields the more focus to the lower values for a better distillation. However, this doesn't hold when the model generates the output with less confidence, or when predicting tasks following multi-modal distributions such as natural language processing~\cite{DBLP:conf/coling/CuiQGZXWLSZL24}. This violation results in that the method with the highest accuracy varies across different tests. Another aspect of this approximation is how to handle the temperature after a student gets the strong correlation against a teacher. In this case, there are less information in the correlation of sample-wise logits. Focusing on other attributes would improve a distillation performance.

\subsection{Analyses}
\begin{figure}[t]
    \centering
    \includegraphics[width=0.8\linewidth]{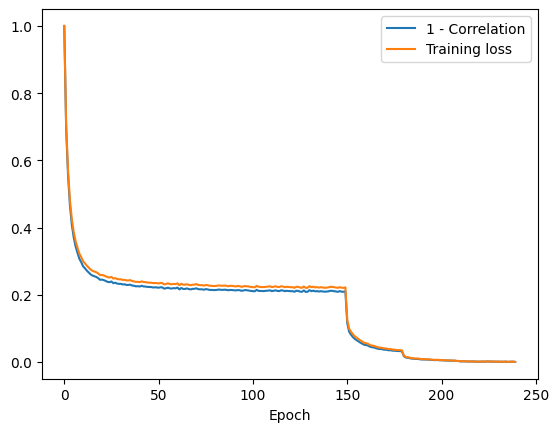}
    \caption{Training Loss and Correlation per Epoch. Note that the curve of the correlation subtracts one for a comparison. A teacher and a student are vgg113 and vgg8, respectively. Note that the correlation of logits in z-score standardized dataset is equivalent to cosine similarity.}
    \label{fig:training_loss_and_correlation}
\end{figure}

\textbf{Correlation} As shown on Fig.~\ref{fig:training_loss_and_correlation}, The correlation of the sample-wise logits aligns with the training loss. Our method help a student resemble for the relation of the logits. The curve supports the assumption of the utilization of either correlation or cosine similarity in previous studies~\cite{DBLP:conf/iccv/PengJLZWLZ019, 10.1609/aaai.v37i2.25236}. In z-score standardized datasets, both are the same.

\begin{figure}[t]
    \centering
    \begin{subfigure}{0.3\textwidth}
        \centering
        \includegraphics[width=\linewidth]{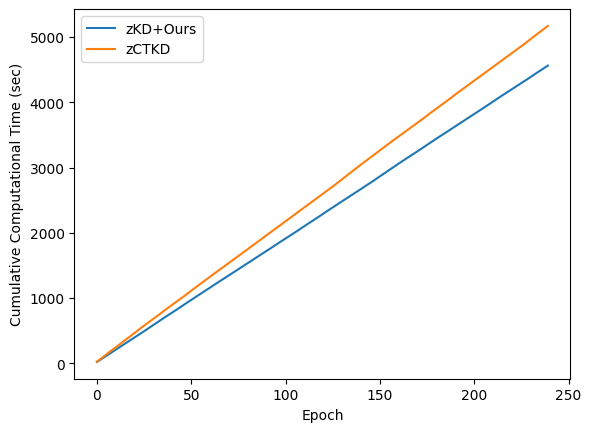}
        \caption{MobileNetV2 \textrightarrow WRN-40-2}
        \label{fig:image1}
    \end{subfigure}%
    \hfill
    \begin{subfigure}{0.3\textwidth}
        \centering
        \includegraphics[width=\linewidth]{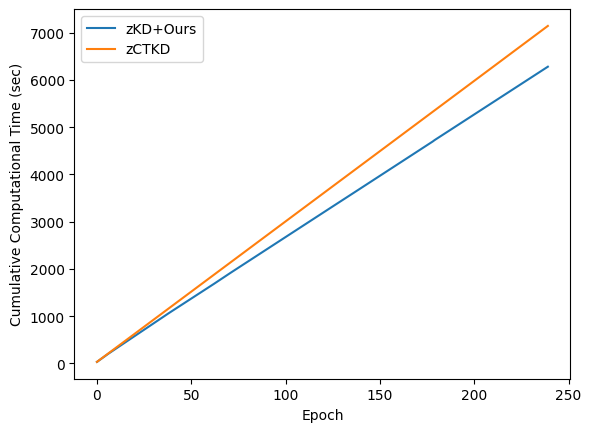}
        \caption{ShuffleV2 \textrightarrow ResNet32x4}
        \label{fig:image2}
    \end{subfigure}
    \hfill
    \begin{subfigure}{0.3\textwidth}
        \centering
        \includegraphics[width=\linewidth]{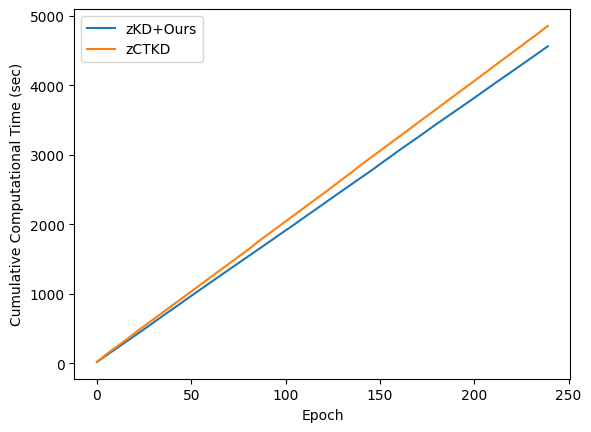}
        \caption{MobileNetV2 \textrightarrow VGG13}
        \label{fig:image3}
    \end{subfigure}
    \caption{The cumulative computational time per epoch with different architecture. Our method determines a temperature faster than other method. The performance is evaluated with NVIDIA V100 GPU.}
    \label{fig:test3_cpu_time}
\end{figure}

\textbf{Computational Time} We demonstrate that our method is faster than previous methods. Fig.~\ref{fig:test3_cpu_time} shows a cumulative computational time per epoch. This performance improvement highlights about 10\% reduction of the total computational time for different architectures. Our method can calculate the adaptive temperature from a teacher in an evaluation mode prior to the distillation process. In this case, This makes the process more efficient as the temperature can be pre-loaded.

\textbf{Ablation Study} We conduct an ablation study to explore the optimal hyper-parameters. Tab.~\ref{tab:test4_ablation} shows how different hyper-parameter settings affect accuracy. For this investigation, the weighted coefficient for Cross Entropy (CE) loss is set to 1 by default. Our method enhances the distillation process when it is assigned a higher weight than the CE loss. As shown in Tab.~\ref{tab:hyperparameter_settings}, this approach reduces the number of parameters that need to be optimized.



%
%
%

\begin{table*}[htbp]
    \centering
    \setlength{\tabcolsep}{2pt}
    \begin{tabular}{llccccccc}
        \toprule
        \multirow{4}{*}{Type} & \multirow{2}{*}{Teacher} & ResNet32x4 & ResNet32x4 & ResNet32x4 & WRN-40-2 & WRN-40-2 & VGG13 & ResNet50 \\
         & & 79.34 & 79.34 & 79.34 & 76.4 & 76.4 & 75.19 & 79.51 \\
         & \multirow{2}{*}{Student} & SHN-V2 & WRN-16-2 & WRN-40-2 & ResNet8x4 & MN-V2 & MN-V2 & MN-V2 \\
         & & 72.87 & 73.49 & 76.4 & 72.42 & 65.53 & 65.53 & 65.53 \\
        \midrule
        \multirow{8}{*}{\textbf{Feature}} & FitNet~\cite{DBLP:journals/corr/RomeroBKCGB14} & 74.15 & 73.97 & 76.75 & 73.3 & 65.66 & 64.87 & 64.59 \\
         & AT~\cite{DBLP:conf/iclr/ZagoruykoK17} & 73.27 & 74.33 & 77.44 & 73.92 & 65.85 & 60.65 & 57.48 \\
         & RKD~\cite{park2019relational} & 73.84 & 73.2 & 76.32 & 72.66 & 65.24 & 65.84 & 65.43 \\
         & CRD~\cite{DBLP:conf/iclr/TianKI20}
 & 76.14 & 75.68 & 78.33 & 75.14 & \textcolor{red}{\textbf{69.91}} & 69.4 & 69.32 \\
         & Review KD~\cite{DBLP:conf/cvpr/Chen0ZJ21} & 74.64 & 73.89 & 77.12 & 73.51 & 68.5 & 65.74 & 63.41 \\
         & SimKD~\cite{DBLP:conf/cvpr/ChenMZWF022} & 76.14 & 75.36 & 78.29 & 74.39 & 67.22 & 67.32 & 67.24 \\
         & CAT-KD~\cite{DBLP:conf/cvpr/GuoYL023} & 72.98 & 73.87 & 76.74 & 72.65 & 65.29 & 65.78 & 64.92 \\
        \midrule
        \multirow{8}{*}{\textbf{Logit}} & TTM~\cite{zheng2024knowledge} & 75.39 & 75.68 & 77.96 & 75.93 & 68.18 & 67.91 & 67.8 \\
         & zKD~\cite{DBLP:journals/corr/HintonVD15, Sun_2024_CVPR} & \textcolor{red}{\textbf{76.62}} & 75.05 & 78.45 & 76.53 & 69.39 & \textcolor{red}{\textbf{69.79}} & 69.21 \\
         & zCTKD~\cite{10.1609/aaai.v37i2.25236, Sun_2024_CVPR} & 76.18 & 75.2 & 77.91 & \textcolor{red}{\textbf{77.01}} & 68.62 & 69.32 & 69.19 \\
         & zKD+Ours & 76.52 & 75.56 & 78.41 & 75.02 & 69.89 & 69.47 & \textcolor{red}{\textbf{70.11}} \\
         & zDKD~\cite{Zhao_2022_CVPR, Sun_2024_CVPR} & 76.41 & 75.72 & \textcolor{red}{\textbf{79.13}} & 74.55 & 68.66 & 69.04 & 69.19 \\
         & zDKD+Ours & 75.96 & \textcolor{red}{\textbf{76.05}} & 78.86 & 74.82 & 68.95 & 68.44 & 69.78 \\
        \bottomrule
    \end{tabular}
    \caption{Comparison of knowledge distillation methods with different student and teacher architectures. The highest accuracy for each column is marked with bold and colored by red. Methods starting with z notates that it uses z-score standardization.}
    \label{tab:test1_diff_arch}
\end{table*}

\begin{table*}[htbp]
    \centering
    \setlength{\tabcolsep}{2pt}
    \begin{tabular}{llccccccc}
        \toprule
        \multirow{4}{*}{Type} & \multirow{2}{*}{Teacher} & WRN-40-2 & WRN-40-2 & ResNet56 & ResNet110 & ResNet110 & ResNet32×4 & VGG13 \\
         & & 76.4 & 76.4 & 72.96 & 74.14 & 74.14 & 79.34 & 75.19 \\
         & \multirow{2}{*}{Student} & WRN-16-2 & WRN-40-1 & ResNet20 & ResNet20 & ResNet8×4 & ResNet32 & VGG8 \\
         & & 73.49 & 72.39 & 70.08 & 70.08 & 71.69 & 72.42 & 70.69 \\
        \midrule
        \multirow{8}{*}{\textbf{Feature}} & FitNet~\cite{DBLP:journals/corr/RomeroBKCGB14} & 73.7 & 71.95 & 69.22 & 68.85 & 70.94 & 73.62 & 71.26 \\
         & AT~\cite{DBLP:conf/iclr/ZagoruykoK17} & 74.57 & 72.78 & 70.45 & 69.9 & 72.44 & 73.89 & 71.49 \\
         & RKD~\cite{park2019relational} & 73.64 & 71.66 & 70.16 & 69.97 & 71.96 & 72.24 & 71.62 \\
         & CRD~\cite{DBLP:conf/iclr/TianKI20}
 & 75.7 & 74.1 & 71.43 & 71.41 & \textcolor{red}{\textbf{73.98}} & 75.6 & 73.49 \\
         & Review KD~\cite{DBLP:conf/cvpr/Chen0ZJ21} & 73.7 & 72.25 & 68.58 & 68.94 & 71.34 & 73.2 & 72.52 \\
         & SimKD~\cite{DBLP:conf/cvpr/ChenMZWF022} & 73.24 & 72.42 & 67.27 & 68.04 & 70.81 & 74.55 & 73.7 \\
         & CAT-KD~\cite{DBLP:conf/cvpr/GuoYL023} & 73.1 & 71.68 & 69.95 & 69.55 & 71.39 & 72.72 & 70.86 \\
        \midrule
        \multirow{8}{*}{\textbf{Logit}} & TTM~\cite{zheng2024knowledge} & 76.25 & 74 & 71.38 & 71.06 & 73.53 & 75.81 & 74.37 \\
         & WTTM~\cite{zheng2024knowledge} & 75.62 & 73.63 & 71.62 & 71.25 & 73.81 & 74.8 & 73.5 \\
         & zKD~\cite{DBLP:journals/corr/HintonVD15, Sun_2024_CVPR} & 75.9 & 74.68 & 71.06 & 71.09 & 73.52 & 76.24 & 74.135 \\
         & zCTKD~\cite{10.1609/aaai.v37i2.25236, Sun_2024_CVPR}
 & \textcolor{red}{\textbf{76.28}} & 74.45 & 71.49 & \textcolor{red}{\textbf{71.83}} & 73.87 & 76.27 & 73.99 \\
         & zKD+Ours & 76.03 & 74.45 & 70.87 & 70.57 & 73.17 & 76.11 & 74.54 \\
         & zDKD~\cite{Zhao_2022_CVPR, Sun_2024_CVPR} & 75.17 & 74.09 & 70.45 & 70.79 & 72.76 & 75.28 & 73.89 \\
         & zDKD+Ours & 75.66 & \textcolor{red}{\textbf{74.85}} & \textcolor{red}{\textbf{71.9}} & 71.22 & 73.24 & \textcolor{red}{\textbf{76.94}} & \textcolor{red}{\textbf{75.12}} \\
        \bottomrule
    \end{tabular}
    \caption{Comparison of knowledge distillation methods with different student and teacher architectures.The highest accuracy for each column is marked with bold and colored by red. Methods starting with z notates that it uses z-score standardization.}
    \label{tab:test2_same_arch}
\end{table*}

\begin{table*}[htbp]
    \centering
    \begin{tabular}{lccccccccccccc}
        \toprule
        & TTM~\cite{zheng2024knowledge} & WTTM~\cite{zheng2024knowledge} & zKD~\cite{DBLP:journals/corr/HintonVD15, Sun_2024_CVPR} & zCTKD~\cite{10.1609/aaai.v37i2.25236, Sun_2024_CVPR} & zKD+Ours & zDKD~\cite{Zhao_2022_CVPR, Sun_2024_CVPR} & zDKD+Ours \\
        \midrule
        $\lambda_{CE}$ & 1 & 1 & 0.1 & 0.1 & 0.1 & 1 & 1\\
        $\lambda_{KD}$ & 1 & 1 & 9 & 9 & 9 & 1 & 9\\
        $\tau$ & 0.2 & 0.2 & 2 & 4 & - & 4 & - \\
        \bottomrule
    \end{tabular}
    \caption{The setting of weighted parameters show different values across methods. TTM and WTTM have the value of ttm coefficient as a variable, instead of $\tau$. Our method doesn't require any initial value of $\tau$}
    \label{tab:hyperparameter_settings}
\end{table*}

\begin{table}[htbp]
    \centering
    \setlength{\tabcolsep}{2pt}
    \begin{tabular}{lccccccc}
        \toprule
        $\lambda_{KD}$ & 0.9 & 3 & 6 & 9 & 12 & 15 & 18 \\
        \midrule
        zKD+Ours & 73.32 & 73.42 & 74.24 & 75.9 & 76.27 & 76.14 & 76.38 \\
        \bottomrule
    \end{tabular}
    \caption{The result of different hyper-parameter $\lambda_{KD}$ as weighted parameter of KD loss. We select ResNet32×4 and ResNet8×4 as the teacher and student, respectively}
    \label{tab:test4_ablation}
\end{table}
\section{Conclusion}
\label{sec:Conclusion}
In this study, we uncover that the approximation of the distillation loss converges the correlation of logits by a teacher and a student. This reinforces the previous argument that the distillation conveys relevance of logits. Following the observation, we propose a novel approach to calculate a temperature. Our method only refers the maximum logit generated by a teacher model, which reduce a computational time against state-of-the-art methods. Our method shows a promising result in different student and teacher models on a standard benchmark dataset. The adaptive temperature help other methods obtain a better accuracy. This approximating algorithm yields a higher temperature compared to the commonly used static values in testing. The application of the adaptive temperature to multi-modal distribution boosts the efficacy to other tasks.

{
    \small
    \bibliographystyle{ieeenat_fullname}
    \bibliography{main}

\begin{thebibliography}{25}
\providecommand{\natexlab}[1]{#1}
\providecommand{\url}[1]{\texttt{#1}}
\expandafter\ifx\csname urlstyle\endcsname\relax
  \providecommand{\doi}[1]{doi: #1}\else
  \providecommand{\doi}{doi: \begingroup \urlstyle{rm}\Url}\fi

\bibitem[Banerjee et~al.(2021)Banerjee, C., Gupta, Vyas, H., and Mishra]{DBLP:conf/delta/0001CGVHM21}
Kunal Banerjee, Vishak~Prasad C., Rishi~Raj Gupta, Kartik Vyas, Anushree H., and Biswajit Mishra.
\newblock Exploring alternatives to softmax function.
\newblock In \emph{Proceedings of the 2nd International Conference on Deep Learning Theory and Applications, DeLTA 2021, Online Streaming, July 7-9, 2021}, pages 81--86. {SCITEPRESS}, 2021.

\bibitem[Chandrasegaran et~al.(2022)Chandrasegaran, Tran, Zhao, and Cheung]{DBLP:conf/icml/ChandrasegaranT22}
Keshigeyan Chandrasegaran, Ngoc{-}Trung Tran, Yunqing Zhao, and Ngai{-}Man Cheung.
\newblock Revisiting label smoothing and knowledge distillation compatibility: What was missing?
\newblock In \emph{International Conference on Machine Learning, {ICML} 2022, 17-23 July 2022, Baltimore, Maryland, {USA}}, pages 2890--2916. {PMLR}, 2022.

\bibitem[Chen et~al.(2022)Chen, Mei, Zhang, Wang, Feng, and Chen]{DBLP:conf/cvpr/ChenMZWF022}
Defang Chen, Jian{-}Ping Mei, Hailin Zhang, Can Wang, Yan Feng, and Chun Chen.
\newblock Knowledge distillation with the reused teacher classifier.
\newblock In \emph{{IEEE/CVF} Conference on Computer Vision and Pattern Recognition, {CVPR} 2022, New Orleans, LA, USA, June 18-24, 2022}, pages 11923--11932. {IEEE}, 2022.

\bibitem[Chen et~al.(2021)Chen, Liu, Zhao, and Jia]{DBLP:conf/cvpr/Chen0ZJ21}
Pengguang Chen, Shu Liu, Hengshuang Zhao, and Jiaya Jia.
\newblock Distilling knowledge via knowledge review.
\newblock In \emph{{IEEE} Conference on Computer Vision and Pattern Recognition, {CVPR} 2021, virtual, June 19-25, 2021}, pages 5008--5017. Computer Vision Foundation / {IEEE}, 2021.

\bibitem[Cui et~al.(2024)Cui, Qin, Gao, Zhang, Xu, Wu, Li, Sun, Zhou, and Li]{DBLP:conf/coling/CuiQGZXWLSZL24}
Xiao Cui, Yulei Qin, Yuting Gao, Enwei Zhang, Zihan Xu, Tong Wu, Ke Li, Xing Sun, Wengang Zhou, and Houqiang Li.
\newblock Sinkhorn distance minimization for knowledge distillation.
\newblock In \emph{Proceedings of the 2024 Joint International Conference on Computational Linguistics, Language Resources and Evaluation, {LREC/COLING} 2024, 20-25 May, 2024, Torino, Italy}, pages 14846--14858. {ELRA} and {ICCL}, 2024.

\bibitem[de~Br{\'{e}}bisson and Vincent(2016)]{DBLP:journals/corr/BrebissonV15}
Alexandre de Br{\'{e}}bisson and Pascal Vincent.
\newblock An exploration of softmax alternatives belonging to the spherical loss family.
\newblock In \emph{4th International Conference on Learning Representations, {ICLR} 2016, San Juan, Puerto Rico, May 2-4, 2016, Conference Track Proceedings}, 2016.

\bibitem[Guo et~al.(2023)Guo, Yan, Li, and Lin]{DBLP:conf/cvpr/GuoYL023}
Ziyao Guo, Haonan Yan, Hui Li, and Xiaodong Lin.
\newblock Class attention transfer based knowledge distillation.
\newblock In \emph{{IEEE/CVF} Conference on Computer Vision and Pattern Recognition, {CVPR} 2023, Vancouver, BC, Canada, June 17-24, 2023}, pages 11868--11877. {IEEE}, 2023.

\bibitem[Hinton et~al.(2015)Hinton, Vinyals, and Dean]{DBLP:journals/corr/HintonVD15}
Geoffrey~E. Hinton, Oriol Vinyals, and Jeffrey Dean.
\newblock Distilling the knowledge in a neural network.
\newblock \emph{CoRR}, abs/1503.02531, 2015.

\bibitem[Hu et~al.(2023)Hu, Li, Liu, Wu, Chen, Wang, and Liu]{DBLP:journals/corr/abs-2308-04268}
Chengming Hu, Xuan Li, Dan Liu, Haolun Wu, Xi Chen, Ju Wang, and Xue Liu.
\newblock Teacher-student architecture for knowledge distillation: {A} survey.
\newblock \emph{CoRR}, abs/2308.04268, 2023.

\bibitem[Jin et~al.(2023)Jin, Wang, and Lin]{Jin_2023_CVPR}
Ying Jin, Jiaqi Wang, and Dahua Lin.
\newblock Multi-level logit distillation.
\newblock In \emph{Proceedings of the IEEE/CVF Conference on Computer Vision and Pattern Recognition (CVPR)}, pages 24276--24285, 2023.

\bibitem[Kaleem et~al.(2024)Kaleem, Rouf, Habib, Saleem, and Lall]{DBLP:journals/corr/abs-2404-00936}
Sheikh~Musa Kaleem, Tufail Rouf, Gousia Habib, Tausifa~Jan Saleem, and Brejesh Lall.
\newblock A comprehensive review of knowledge distillation in computer vision.
\newblock \emph{CoRR}, abs/2404.00936, 2024.

\bibitem[Krizhevsky et~al.(2009)Krizhevsky, Hinton, et~al.]{krizhevsky2009learning}
Alex Krizhevsky, Geoffrey Hinton, et~al.
\newblock Learning multiple layers of features from tiny images.
\newblock 2009.

\bibitem[Li et~al.(2023)Li, Li, Yang, Zhao, Song, Luo, Li, and Yang]{10.1609/aaai.v37i2.25236}
Zheng Li, Xiang Li, Lingfeng Yang, Borui Zhao, Renjie Song, Lei Luo, Jun Li, and Jian Yang.
\newblock Curriculum temperature for knowledge distillation.
\newblock In \emph{Proceedings of the Thirty-Seventh AAAI Conference on Artificial Intelligence and Thirty-Fifth Conference on Innovative Applications of Artificial Intelligence and Thirteenth Symposium on Educational Advances in Artificial Intelligence}. AAAI Press, 2023.

\bibitem[Liu et~al.(2022{\natexlab{a}})Liu, Liu, Li, and Liu]{DBLP:journals/corr/abs-2202-07940}
Jihao Liu, Boxiao Liu, Hongsheng Li, and Yu Liu.
\newblock Meta knowledge distillation.
\newblock \emph{CoRR}, abs/2202.07940, 2022{\natexlab{a}}.

\bibitem[Liu et~al.(2022{\natexlab{b}})Liu, Liu, Li, and Liu]{liu2022metaknowledgedistillation}
Jihao Liu, Boxiao Liu, Hongsheng Li, and Yu Liu.
\newblock Meta knowledge distillation, 2022{\natexlab{b}}.

\bibitem[Park et~al.(2019)Park, Kim, Lu, and Cho]{park2019relational}
Wonpyo Park, Dongju Kim, Yan Lu, and Minsu Cho.
\newblock Relational knowledge distillation.
\newblock In \emph{Proceedings of the IEEE Conference on Computer Vision and Pattern Recognition}, pages 3967--3976, 2019.

\bibitem[Peng et~al.(2019)Peng, Jin, Li, Zhou, Wu, Liu, Zhang, and Liu]{DBLP:conf/iccv/PengJLZWLZ019}
Baoyun Peng, Xiao Jin, Dongsheng Li, Shunfeng Zhou, Yichao Wu, Jiaheng Liu, Zhaoning Zhang, and Yu Liu.
\newblock Correlation congruence for knowledge distillation.
\newblock In \emph{2019 {IEEE/CVF} International Conference on Computer Vision, {ICCV} 2019, Seoul, Korea (South), October 27 - November 2, 2019}, pages 5006--5015. {IEEE}, 2019.

\bibitem[Romero et~al.(2015)Romero, Ballas, Kahou, Chassang, Gatta, and Bengio]{DBLP:journals/corr/RomeroBKCGB14}
Adriana Romero, Nicolas Ballas, Samira~Ebrahimi Kahou, Antoine Chassang, Carlo Gatta, and Yoshua Bengio.
\newblock Fitnets: Hints for thin deep nets.
\newblock In \emph{3rd International Conference on Learning Representations, {ICLR} 2015, San Diego, CA, USA, May 7-9, 2015, Conference Track Proceedings}, 2015.

\bibitem[Sun et~al.(2024)Sun, Ren, Li, Wang, and Cao]{Sun_2024_CVPR}
Shangquan Sun, Wenqi Ren, Jingzhi Li, Rui Wang, and Xiaochun Cao.
\newblock Logit standardization in knowledge distillation.
\newblock In \emph{Proceedings of the IEEE/CVF Conference on Computer Vision and Pattern Recognition (CVPR)}, pages 15731--15740, 2024.

\bibitem[Tian et~al.(2020)Tian, Krishnan, and Isola]{DBLP:conf/iclr/TianKI20}
Yonglong Tian, Dilip Krishnan, and Phillip Isola.
\newblock Contrastive representation distillation.
\newblock In \emph{8th International Conference on Learning Representations, {ICLR} 2020, Addis Ababa, Ethiopia, April 26-30, 2020}. OpenReview.net, 2020.

\bibitem[Vincent et~al.(2015)Vincent, de~Br{\'{e}}bisson, and Bouthillier]{DBLP:conf/nips/VincentBB15}
Pascal Vincent, Alexandre de Br{\'{e}}bisson, and Xavier Bouthillier.
\newblock Efficient exact gradient update for training deep networks with very large sparse targets.
\newblock In \emph{Advances in Neural Information Processing Systems 28: Annual Conference on Neural Information Processing Systems 2015, December 7-12, 2015, Montreal, Quebec, Canada}, pages 1108--1116, 2015.

\bibitem[Yang et~al.(2024)Yang, Lu, Zhu, Wang, Chen, Gao, Yan, and Chen]{DBLP:journals/corr/abs-2407-01885}
Chuanpeng Yang, Wang Lu, Yao Zhu, Yidong Wang, Qian Chen, Chenlong Gao, Bingjie Yan, and Yiqiang Chen.
\newblock Survey on knowledge distillation for large language models: Methods, evaluation, and application.
\newblock \emph{CoRR}, abs/2407.01885, 2024.

\bibitem[Zagoruyko and Komodakis(2017)]{DBLP:conf/iclr/ZagoruykoK17}
Sergey Zagoruyko and Nikos Komodakis.
\newblock Paying more attention to attention: Improving the performance of convolutional neural networks via attention transfer.
\newblock In \emph{5th International Conference on Learning Representations, {ICLR} 2017, Toulon, France, April 24-26, 2017, Conference Track Proceedings}. OpenReview.net, 2017.

\bibitem[Zhao et~al.(2022)Zhao, Cui, Song, Qiu, and Liang]{Zhao_2022_CVPR}
Borui Zhao, Quan Cui, Renjie Song, Yiyu Qiu, and Jiajun Liang.
\newblock Decoupled knowledge distillation.
\newblock In \emph{Proceedings of the IEEE/CVF Conference on Computer Vision and Pattern Recognition (CVPR)}, pages 11953--11962, 2022.

\bibitem[Zheng and Yang(2024)]{zheng2024knowledge}
Kaixiang Zheng and En-Hui Yang.
\newblock Knowledge distillation based on transformed teacher matching.
\newblock In \emph{The Twelfth International Conference on Learning Representations (ICLR 2024)}, 2024.

\end{thebibliography}
}


\end{document}